\documentclass[conference]{IEEEtran}
\usepackage{spconf,amsmath,epsfig}

\usepackage{simpleConference}
\usepackage{url}

\usepackage[table]{xcolor}
\usepackage{multirow}
\usepackage{booktabs}
\usepackage{array}
\usepackage{amssymb}
\usepackage[colorlinks,linkcolor=red]{hyperref}

\let\OLDthebibliography\thebibliography
\renewcommand\thebibliography[1]{
  \OLDthebibliography{#1}
  \setlength{\parskip}{0pt}
  \setlength{\itemsep}{0pt plus 0.3ex}
}

\begin{document}\sloppy

\def\x{{\mathbf x}}
\def\L{{\cal L}}

\title{SimViT: Exploring a Simple Vision Transformer with sliding windows}

\author{Gang Li$^{1,3,*}$; Di Xu$^{2,3,*}$; Xing Cheng$^{2,3}$; Lingyu Si$^{1,3}$; Changwen Zheng$^{1}$ \\ ucasligang@gmail.com; chengxing20s@ict.ac.cn; hanielxx@outlook.com; \\ lingyu@iscas.ac.cn; changwen@iscas.ac.cn 
\\ \\
Institute of Software, Chinese Academy of Science$^1$; \\ Institute of Computing Technology, Chinese Academy of Sciences$^2$; \\ University of Chinese Academy of Sciences$^3$
}

\maketitle

\begin{abstract}

Although vision Transformers have achieved excellent performance as backbone models in many vision tasks, most of them intend to capture global relations of all tokens in an image or a window, which disrupts the inherent spatial and local correlations between patches in 2D structure. In this paper, we introduce a simple vision Transformer named SimViT, to incorporate spatial structure and local information into the vision Transformers. Specifically, we introduce Multi-head Central Self-Attention(MCSA) instead of conventional Multi-head Self-Attention to capture highly local relations. The introduction of sliding windows facilitates the capture of spatial structure.  Meanwhile, SimViT extracts multi-scale hierarchical features from different layers for dense prediction tasks. Extensive experiments show the SimViT is effective and efficient as a general-purpose backbone model for various image processing tasks. Especially, our SimViT-Micro only needs 3.3M parameters to achieve 71.1\% top-1 accuracy on ImageNet-1k dataset, which is the smallest size vision Transformer model by now. Our code will be available in \url{https://github.com/ucasligang/SimViT}.

\end{abstract}
\section{Introduction}

\begin{figure}[t]
  \begin{center}
    \includegraphics[width=3.5in]{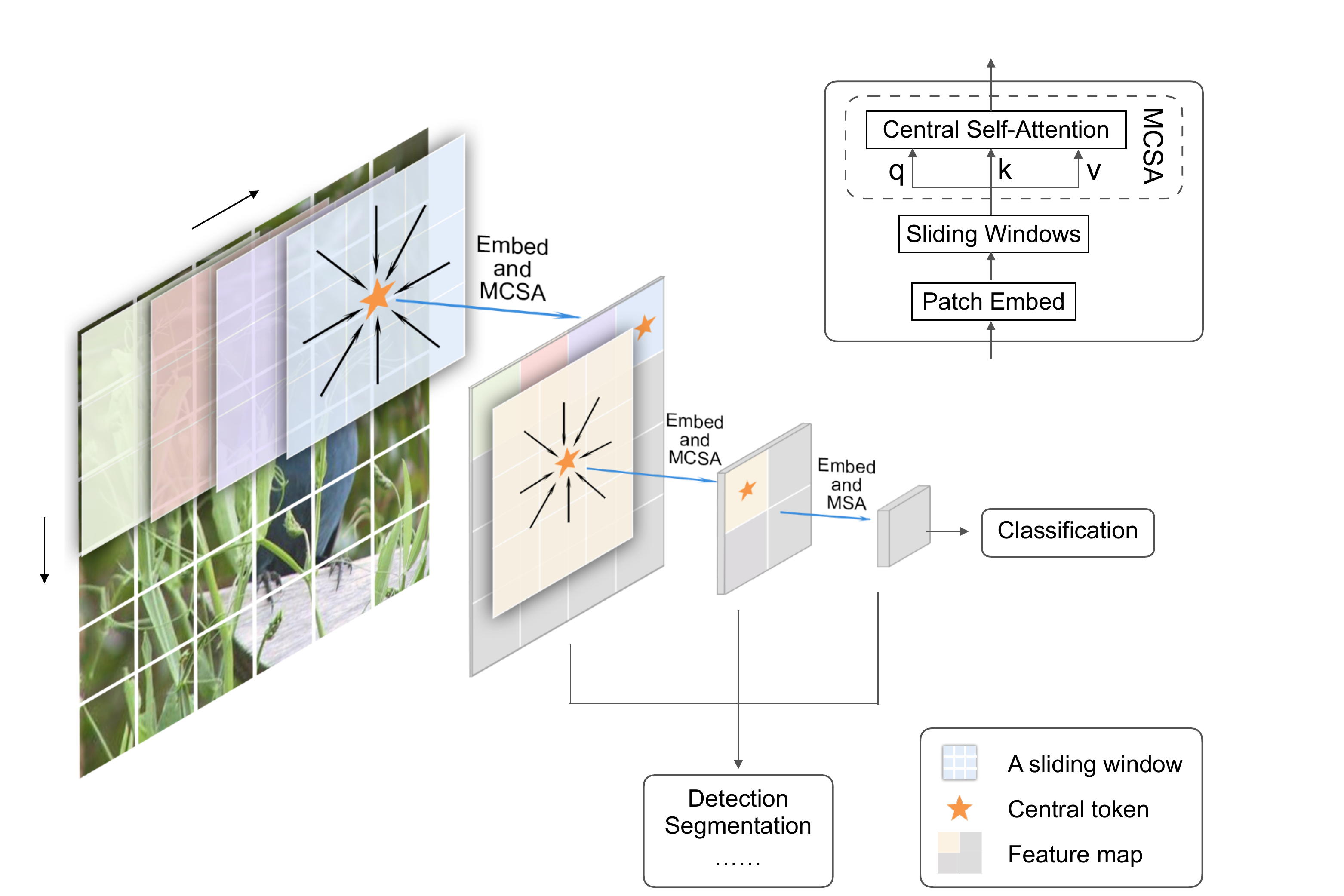}
  \end{center}

  \caption{\small The overview of our SimViT. An image is split into many patches. The proposed Multi-head Central Self-Attention(MCSA) takes features of patches as input tokens, where the feature of the central patch is used as the query and features of surrounding patches are used as keys and values in each window. The sliding windows are used to capture visual spatial structure. Multi-scale features are extracted for different tasks(classification, detection, segmentation). }
  \label{fig:motivation}
  \vspace{-0.5cm}
\end{figure}

\label{sec:intro}
  Recent studies\cite{Swin,PVT} are devoted to propagating the advances of Transformers into computer vision and have made rapid progress. To employ the self-attention blocks originally designed for sequence data, most existing vision Transformers forcibly convert image data into sequences by flattening 2D patches as 1D tokens. However, the inherent spatial information and highly local correlations among patches in 2D structures are broken. Although this can be alleviated by introducing position embeddings\cite{Swin}, it brings additional learnable parameters. Moreover, translation invariant is lacking for present vision Transformers because of the uniqueness of positional encodings.


On the other hand, CNNs have proved the importance of preserving the highly local relations with regional  connections in convolution kernels. Convolutions with sliding windows naturally capture the 2D spatial information without additional position encodings. Meanwhile, the convolutional operation is invariant to translations because the computation of responses is only related to shared weights and  ``pixels" within the same window. 




Inspired by the practices from CNNs, we seek for incorporating spatial structure and local information into vision Transformer with simple sliding windows. In this paper, we propose a simple yet effective vision Transformer named SimViT, as shown in Figure \ref{fig:motivation}. Specifically, We introduce overlapping sliding windows for aggregating spatial information and cross-window connection. In the previous layers, Multi-head Central Self-Attention (MCSA) is employed to replace the standard Multi-head Self-Attention (MSA) within each window. MCSA computes the attention map by taking the feature vector of the central patch as query and the features of surrounding patches as keys and values. Only the response of the central patch is calculated to capture the highly local relations and reduce the computational complexity. In later layers, conventional MSA is applied to establish global dependencies. In this way, our SimViT naturally integrate local and global visual dependencies, further enhancing the capacity of modeling. The output of our SimViT is only related to local and global context without additional unique positional encodings, which brings translation invariance. Moreover, the multi-scale hierarchical structure is introduced into our SimViT, which is important for dense prediction tasks.



By taking both advantages of local structure-preserving with sliding window-based hierarchical  architecture in CNNs and information aggregation with self-attention in Transformers,  our work bridges the cognitive gap of CNNs and Transformers for visual data modeling. Our proposed SimViT can serve as a general-purpose backbone model and be applied in various vision tasks. The results of extensive experiments show that SimViT achieves excellent performances in different vision tasks, including image classification, object detection, semantic segmentation. It is worth mentioning that the Micro version of our SimViT achieves 71.1\% top-1 accuracy on ImageNet-1k image classification only needs 3.3M parameters. To our knowledge, SimViT-Micro is the smallest size vision Transformer model by now.

Our main contributions are fourfold: (1) We propose a simple yet effective vision Transformer named SimViT to incorporate spatial structure and cross-window connection with sliding windows into  vision  Transformer; (2) We introduce a new attention module, MCSA instead of the standard MSA to build highly local relations in a window;
(3) Our work bridges the cognitive gap of CNNs and Transformers for visual data modeling, which can encourage more previous practices of CNNs applied to vision Transformers; (4) Our models show excellent performance in different vision tasks, like image classification, object detection, semantic segmentation.

\section{Related Work}
\textbf{Self-attention in CNNs.} Convolutional Neural Networks (CNNs) have been dominant in the field of computer vision for a long time. With the popularity of the self-attention mechanism in recent years, researchers have begun to combine CNNs with self-attention. The non-local operation\cite{NonLocalNetwork} is proposed for capturing long-term connections in spatial structure. 
SENet\cite{SENets} proposes the channel-wise attention to model inter-dependencies between channels, CBAM\cite{Cbam} further aggregates the spatial-wise attention and channel-wise attention. In image classification tasks, Local Relation Network(LR-Net)\cite{Hu2019LocalRN} presents a relational extractor that adaptively determines the weights of local pixel pairs to composite low-level visual elements into higher-level entities. Based on LR-Net\cite{Hu2019LocalRN}, the object relation module\cite{Hu2018RelationNF} further equips self-attention to model the relations between the appearance features of objects and the geometry after convolutional modules for object detection. Moreover,  a recent study\cite{StandAloneSlefAttention} intends to explore the possibility of applying self-attention to replace convolutional operators. Our MCSA is similar with \cite{StandAloneSlefAttention}, which proposes spatial-relative attention using 2D relative position embeddings. While our MCSA is integrated into the vision Transformer blocks without explicit position encodings.\\


\textbf{Multi-scale vision Transformers.} Recent works have successfully extended uniform-scale Transformer\cite{ViT} to multi-scale vision Transformers\cite{PVT,PVTv2,Swin} to capture hierarchy features for dense prediction tasks. PVT\cite{PVT} introduces a unified pyramid structure and Spatial-Reduction Attention(SRA) for dense prediction tasks. Overlapping patch embedding, depth-wise convolution, and linear SRA are introduced into PVT v2\cite{PVTv2} to improve the original PVT\cite{PVT}. Swin Transformer\cite{Swin} proposes shifting windows into Transformer blocks. In order to build cross-window connections, two successive Swin Transformer blocks are used to replace one conventional vision Transformer block. Our work only computes self-attention between the central patch and surrounding patches along with simple sliding windows. In this way, our SimViT naturally maintains the highly local relations and spatial structure.


\section{Method}

\subsection{Model architecture}

\begin{figure*}[t]
	\centering
	\includegraphics[width=.9\textwidth]{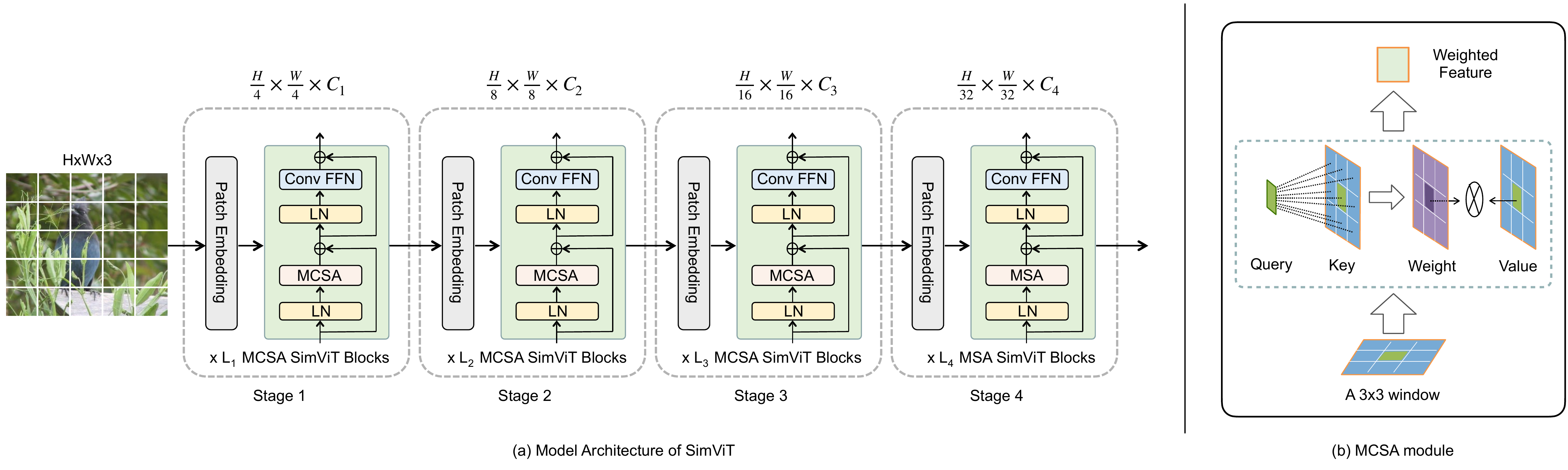}
	\caption{\small (a) The architecture of our SimViT. SimViT includes four stages. Each of the first three stages consists of a patch embedding layer and several MCSA(Multi-head Central Self-Attention) blocks to capture local relations. In the last stage, MCSA is replaced by MSA(Multi-head Self-Attention) to build global connections. (b) Illustration of an MCSA module.}
	\label{fig:structure}
	\vspace{-0.5cm}
\end{figure*}
In order to better tackle different dense prediction tasks such as object detection and semantic segmentation, we introduce a sliding window-based hierarchical structure into our model to generate multi-scale representations. The architecture design of our SimViT model follows~\cite{PVT,PVTv2} and an overview is presented in Figure \ref{fig:structure}(a). SimViT consists of four stages, each stage has a similar structure composed of a patch embedding layer and $L_{i}$ sequentially stacked MCSA/MSA Transformer blocks. The first three stages employ MCSA to build local relations of patches and extract low-level features. MSA is used in our Transformer blocks of the last stage to enhance global connections between patches. Our SimViT integrates local and global visual dependencies to enhance the capacity of modeling.

Following previous works\cite{Swin,PVT}, we split each $H \times W$ RGB image into non-overlapping patches in the first stage, where the size of each patch is $4\times4\times3$.  In the next stages, $2\times2\times C_{i-1}$ non-overlapping patches are divided from previous token map. In the $i$-th stage, patches are mapped into 1$\times C_{i}$-dimensional features by a FC(fully-connected) layer with Layer Normalization. As a result, the output features from four stages are $F_{1}\in \mathbb{R}^{\frac{H}{4}\times\frac{W}{4}\times C_{1}}$, $F_{2}\in \mathbb{R}^{\frac{H}{8}\times\frac{W}{8}\times C_{2}}$, $F_{3}\in \mathbb{R}^{\frac{H}{16}\times\frac{W}{16}\times C_{3}}$, $F_{4}\in \mathbb{R}^{\frac{H}{32}\times\frac{W}{32}\times C_{4}}$. Note that MCSA/MSA SimViT blocks don't change the resolution of feature map in each stage. Multi-scale feature maps \{$F_{1}, F_{2}, F_{3}, F_{4}$\} from four stages are extracted for various dense prediction tasks. 

\textbf{MCSA/MSA SimViT block.} In MCSA/MSA SimViT blocks, we introduce Convolutional Feed Forward Network(Convolutional FFN) from PVT v2\cite{PVTv2}, which is implemented by adding a depth-wise convolution between the first FC(fully-connected) layer and GELU\cite{GELU} in the feed-forward network of conventional Transformer block. Layer Normalization\cite{LayerNormalization} is applied before each MCSA/MSA module and each Convolutional FFN. Residual connection is applied in MCSA/MSA module and Convolutional FFN module. The MCSA/MSA SimViT block is calculated as:

\begin{equation}
    \tilde{h}^{l}=MCSA/MSA(LN(h^{l-1}))+h^{l-1}
    \label{eq:MCSA1}
\end{equation}
\begin{equation}
    h^{l}=ConvFFN(LN(\tilde{h}^{l}))+\tilde{h}^{l}
    \label{eq:MCSA2}
\end{equation}
where  $LN$ is the operation of Layer Normalization, $ConvFFN$ and $MCSA/MSA$ stand for Convolutional FFN and MCSA/MSA module, respectively. $\tilde{h}^{l}$ and $h^{l}$ denote the output feature of MCSA/MSA module and Convolutional FFN module of SimViT block $l$, respectively. $h^{l-1}$ is the output feature of Convolutional FFN module of SimViT block $l-1$.

\subsection{MCSA using sliding windows}

For most vision Transformers, conventional MSA jointly computes pairwise full attentions of flattening tokens, which disrupts the inherent spatial and local correlations between patches in 2D structure. In this section, we introduce the sliding windows scheme into SimViT to capture the spatial structure and cross-window connection. Within each window, we propose MCSA to establish the highly local dependencies between the central patch and surrounding patches.

\begin{figure}[t]
	\includegraphics[width=.5\textwidth]{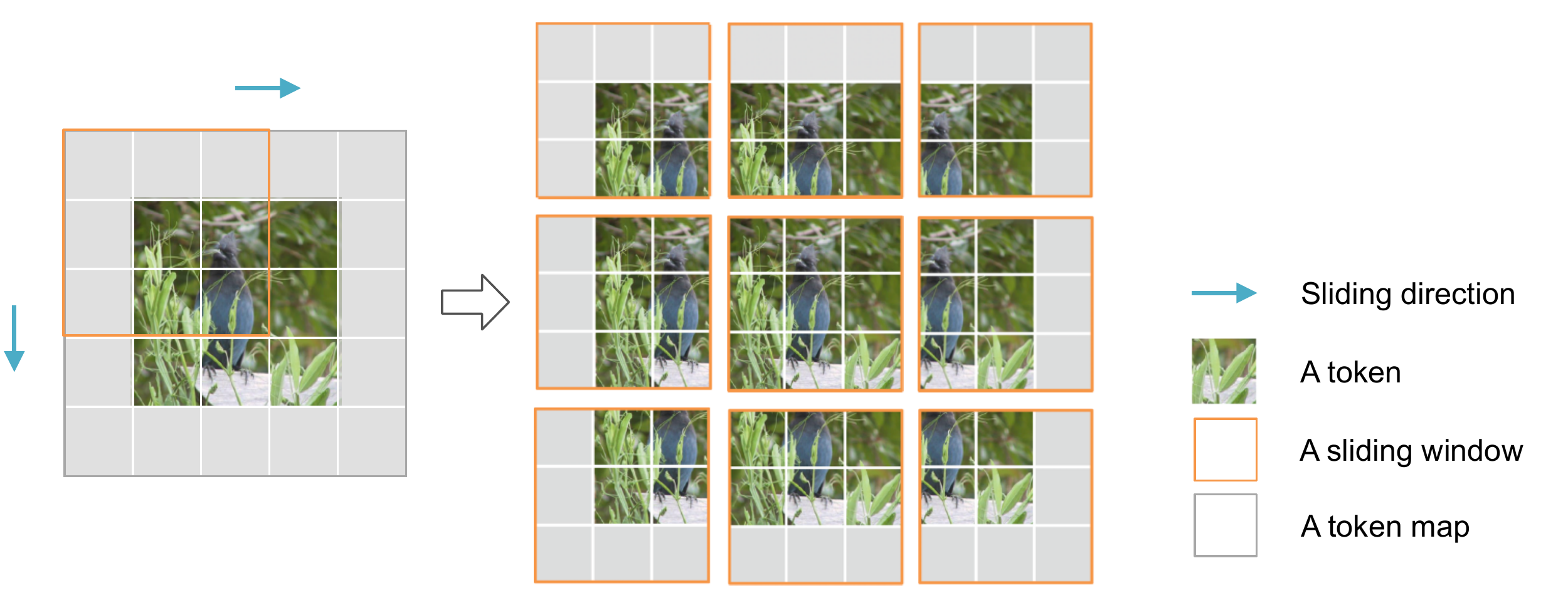}
	\caption{\small An illustration of the sliding window approach. For convenience, we use a 3$\times$ 3 token map and 3x3 sliding windows(the stride size of sliding is 1) for display. Gray tokens represent the zero paddings of the token map. New overlapping windows are generated by the sliding of windows. For the self-attention computation of each new window please refer to \ref{fig:structure}(b).}
	\label{fig:slidingwindows}
	\vspace{-0.5cm}
\end{figure}

\textbf{Cross-window connection with sliding windows.}
We show a simple illustration of the sliding window approach in Figure \ref{fig:slidingwindows}.  We apply overlapping sliding windows to establish cross-window dependency. Formally, given a 2D output token map $x_{i} \in \mathbb{R}^{H_{i} \times W_{i} \times C_{i}}$ from Patch Embedding layer or previous MCSA SimViT block in Stage $i$. The number of overlapping windows is $H_{i}^{'} \times W_{i}^{'}$, $H_{i}^{'} $ and $W_{i}^{'}$ can be formulated as:
\begin{equation}
    H_{i}^{'}=\left\lfloor\frac{H_{i}+2 p-k}{s}\right\rfloor+1,
    W_{i}^{'}=\left\lfloor\frac{W_{i}+2 p-k}{s}\right\rfloor+1
\end{equation}
where $p$ is padding size, $k$ is the size of the sliding window, $s$ is the stride size of the sliding window, respectively.
In our implementation, we set $p$ = 1 and $s$=1 for $3\times3$ sliding windows, which easily keeps the number of tokens equal to the number of overlapping windows for the computation of the central self-attention map.


\begin{table*}
\caption{\small Five model configurations for our SimViT with respect to different modeling capacities are introduced, including SimViT-Micro, SimViT-Tiny, SimViT-Small, SimViT-Medium and SimViT-Large. $P_{i}$, $C_{i}$, $N_{i}$, $E_{i}$ stand for the patch size, the channel number of the output, the head number of self-attention modules and the expansion ratio of the Convolutional FFN\cite{PVTv2} in Stage $i$, respectively.} 
\centering
\newcommand{\tabincell}[2]{\begin{tabular}{@{}#1@{}}#2\end{tabular}}
\renewcommand\arraystretch{1.35} 

\setlength{\tabcolsep}{0.4mm}{
\begin{tabular}{c|c|c|c|c|c|c|c}
\hline
& Output Size &  Layer Name& SimViT-Micro & SimViT-Tiny & SimViT-Small& SimViT-Medium & SimViT-Large\\ \hline
\multirow{3}{*}{Stage 1} & 56$\times$56 &  Patch Embedding  &   $P_{1}=4;C_{1}=32$ &  \multicolumn{4}{c}{$P_{1}=4;C_{1}=64$} \\ \cline{2-8} & \multirow{2}{*}{$56\times56$} & 
\multirow{2}{*}{\tabincell{c}{Transformer \\[-3pt] Encoder}} & \multirow{2}{*}{$\left[
\begin{aligned}N_{1}=1 \\[-3pt] E_{1}=8\end{aligned}
\right]
\times$ 2} & \multirow{2}{*}{$\left[\begin{aligned} N_{1}=1 \\[-3pt] E_{1}=8\end{aligned}\right]\times$ 2} & \multirow{2}{*}{$\left[\begin{aligned}N_{1}=1 \\[-3pt] E_{1}=8 \end{aligned}\right]\times$ 3} & \multirow{2}{*}{$\left[
\begin{aligned}N_{1}=1 \\[-3pt] E_{1}=8\end{aligned}
\right]\times$ 3}& \multirow{2}{*}{$\left[
\begin{aligned}N_{1}=1 \\[-3pt] E_{1}=4\end{aligned}
\right]\times$ 3} \\ &  &  &    &      &      &  \\ \hline
\multirow{3}{*}{Stage 2} & 28$\times$28 &  Patch Embedding  &  $P_{2}=2;C_{2}=64$ &  \multicolumn{4}{c}{$P_{2}=2;C_{2}=128$} \\ \cline{2-8} & \multirow{2}{*}{28$\times$28} & 
\multirow{2}{*}{\tabincell{c}{Transformer \\[-3pt] Encoder}} & \multirow{2}{*}{$\left[
\begin{aligned}N_{2}=2 \\[-3pt] E_{2}=8\end{aligned}
\right]\times$ 3} & \multirow{2}{*}{$\left[\begin{aligned}N_{2}=2 \\[-3pt] E_{2}=8\end{aligned}
\right]\times$ 4} & \multirow{2}{*}{$\left[\begin{aligned}N_{2}=2 \\[-3pt] E_{2}=8\end{aligned}
\right]\times$ 6} & \multirow{2}{*}{$\left[
\begin{aligned}N_{2}=2 \\[-3pt] E_{2}=8\end{aligned}
\right]\times$ 8}& \multirow{2}{*}{$\left[
\begin{aligned}N_{2}=2 \\[-3pt] E_{2}=4\end{aligned}
\right]\times$ 8} \\ &  &  &    &      &      &  \\ \hline
\multirow{3}{*}{Stage 3} & 14$\times$14 &  Patch Embedding  &  $P_{3}=2;C_{3}=160$ &  \multicolumn{4}{c}{$P_{3}=2;C_{3}=320$} \\ \cline{2-8} & \multirow{2}{*}{14$\times$14} & 
\multirow{2}{*}{\tabincell{c}{Transformer \\[-3pt] Encoder}} & \multirow{2}{*}{$\left[
\begin{aligned}N_{3}=5 \\[-3pt] E_{3}=4\end{aligned}
\right]\times$ 3} & \multirow{2}{*}{$\left[\begin{aligned}N_{3}=5 \\[-3pt] E_{3}=4\end{aligned}
\right]\times$ 3} & \multirow{2}{*}{$\left[\begin{aligned}N_{3}=5 \\[-3pt] E_{3}=4\end{aligned}
\right]\times$ 13}  & \multirow{2}{*}{$\left[\begin{aligned}N_{3}=5 \\[-3pt] E_{3}=4\end{aligned}
\right]\times$ 30} & \multirow{2}{*}{$\left[\begin{aligned}N_{3}=5 \\[-3pt] E_{3}=4\end{aligned}
\right]\times$ 40} \\ &  &  &    &      &      &  \\ \hline
\multirow{3}{*}{Stage 4} & 7$\times$7 &  Patch Embedding  &  $P_{4}=2;C_{4}=256$ &  \multicolumn{4}{c}{$P_{4}=2;C_{4}=512$} \\ \cline{2-8} & \multirow{2}{*}{7$\times$7} & 
\multirow{2}{*}{\tabincell{c}{Transformer \\[-3pt]  Encoder}} & \multirow{2}{*}{$\left[\begin{aligned}N_{4}=8 \\[-3pt] E_{4}=4\end{aligned}
\right]\times$ 2} & \multirow{2}{*}{$\left[\begin{aligned}N_{4}=8 \\[-3pt] E_{4}=4\end{aligned}
\right]\times$ 2} & \multirow{2}{*}{$\left[\begin{aligned}N_{4}=8 \\[-3pt] E_{4}=4\end{aligned}
\right]\times$ 3} & \multirow{2}{*}{$\left[\begin{aligned}N_{4}=8 \\[-3pt] E_{4}=4\end{aligned}
\right]\times$ 3}& \multirow{2}{*}{$\left[\begin{aligned}N_{4}=8 \\[-3pt] E_{4}=4\end{aligned}
\right]\times$ 3} \\ &  &  &    &      &      &  \\ \hline
\end{tabular}}
\label{table:ModelConfiguration}
\vspace{-0.5cm}
\end{table*}

\textbf{Multi-head Central Self-Attention.} An illustration of our MCSA is shown in Figure \ref{fig:structure} (b). In order to capture highly local dependencies, the central patch is used as the query to compute the attention map with surrounding patches in a window. The computation of conventional Self-Attention(SA) in MSA as:
\begin{equation}
    SA(Q,K,V)=softmax(\frac{QK^{T}}{\sqrt{d}}+B)V
    \label{eq:MSA}
\end{equation}
where $Q, K, V \in \mathbb{R}^{HW \times d}$ are the query, key, value matrices; $HW$ and $d$ are the number of tokens and the token dimension, respectively; $B$ denotes the bias of positional embedding. While the computation of Central Self-Attention(CSA) in our MCSA as:
\begin{equation}
    CSA(Q,K,V)=softmax(\frac{qK^{T}}{\sqrt{d}})^{T}V,
    \label{eq:MCSA}
\end{equation}
where $q \in \mathbb{R}^{1 \times d}$ denotes the query vector of central token in the 2D token map, other notations are consistent with Equation \ref{eq:MSA}. Similar to MSA, we also introduce Multi-head mechanism to enhance the modeling ability of our MCSA:
\begin{equation}
    MCSA(Q,K,V)=Concat(head_{1},...head_{h})W^{O},
    \label{eq:Multi-head}
\end{equation}
  where $head_{i}=CSA(QW_{i}^{Q},KW_{i}^{K},VW_{i}^{V})$, the projections are parameter matrices $W_{i}^{Q} \in \mathbb{R}^{d\times \hat{d}}$, $W_{i}^{K} \in \mathbb{R}^{d\times \hat{d}}$, $W_{i}^{V} \in \mathbb{R}^{d\times \hat{d}}$, $h$ is the number of projection head, $d$ is the dimension of token, $\hat{d}=d/h$ is output dimension of each projection head. The computation of multi-head attention can easily be completed through the single matrix multiplication.

\subsection{Model configuration}

We consider five different model configurations for our SimViT with different hyper-parameters. Specifically, we describe our model as -Micro, -Tiny, -Small, -Medium and -Large according to the number of parameters, respectively. We show all details of these configures in Table \ref{table:ModelConfiguration}, where the following hyper-parameters are listed:
We design fewer layers in the first stage and the last stage refer to\cite{ResNet,PVT}, and put more computation resources in the middle two stages. Moreover, the long-narrow structure is also introduced into our SimViT following the design principles of conventional CNNs\cite{ResNet}, which decreases spatial dimensions and increases the channel dimension of the feature map stage by stage.



\section{Experiments}
We compare our SimViT with ResNet\cite{ResNet} and recent vision Transformer backbones\cite{PVT,PVTv2,Swin} on three computer vision tasks, image classification, object detection and semantic segmentation.

\subsection{Classification}
We conduct the experiments of image classification on ImageNet-1k\cite{ImagNet},  which contains around 1.28M training images and 50K validation images from 1000 classes. We report the top-1 classification accuracy of our models on the validation set. Following PVT\cite{PVT}, we apply the same data augmentations like random cropping, random horizontal flipping, label-smoothing regularization, and random erasing on the training phase. For pair comparisons, all models are trained for 300 epochs from scratch on 8 V100 GPUs follow\cite{Swin, PVT, PVTv2}. We report our experimental results compared with SOTA methods in Table \ref{table:ImageClassificationCompareSOTA}. All these models achieve excellent performance compared with the state-of-the-art methods. It is noteworthy that the micro version of our SimViT only with 3.3M parameters achieves 71.1\% top-1 classification accuracy on the validation set of ImageNet-1k.

\begin{table}[!h]
\caption{\small The performance of image classification on the ImageNet-1k validation set compared with ResNet\cite{ResNet} and current SOTA methods.} 
\vspace{0.1cm}
\centering
\setlength{\tabcolsep}{0.3mm}{ 
\begin{tabular}{lccc}
\hline
 Method  &  Param(M) &   GFLOPs    &  Top-1 Acc(\%)     \\ \hline
        PVTv2-B0\cite{PVTv2}  &   3.4   & 0.6   &   70.5 \\ 
        \rowcolor{gray!20}  SimViT-Micro  &    3.3   &  0.7  &   \textbf{71.1}    \\ \hline
      ResNet18\cite{ResNet}  &  11.7   & 1.8  & 69.8      \\ 
      PVT-Tiny\cite{PVT}  &   13.2    & 1.9  & 75.1  \\  
      PVTv2-B1\cite{PVT}  &   13.1    & 2.1  & 78.7  \\  
\rowcolor{gray!20}  SimViT-Tiny  &   13.0    &  2.5 &  \textbf{79.3} \\ \hline
  ResNet50\cite{ResNet}    &  25.6  &    4.1   & 76.1   \\ 
  PVT-Small\cite{PVT}  &  24.5 & 3.8  & 79.8  \\
  TNT-S\cite{PVT}  &  23.8 & 5.2  & 81.3  \\
  PVTv2-B2\cite{PVT} &25.4 &4.0& 82.0\\
  Swin-T\cite{Swin}& 29.0 &4.5& 81.3\\
\rowcolor{gray!20}  SimViT-Small  &  29.4  & 6.2 & \textbf{82.6} \\ \hline
 ResNet101\cite{ResNet}    & 44.7 & 7.9 & 77.4  \\ 
      PVT-Medium\cite{PVT} & 44.2 & 6.7 & 81.2\\
      Swin-S\cite{Swin} & 50.0 & 8.7 & 83.0\\
\rowcolor{gray!20}  SimViT-Medium & 51.3 & 10.9  & \textbf{83.3}  \\ \hline
ResNet152\cite{ResNet} & 60.2 & 11.6 & 78.3 \\
PVT-Large\cite{PVT} & 61.4 & 9.8 & 81.7\\
TNT-B\cite{TNT} & 66.0 & 14.1 & 82.8\\
\rowcolor{gray!20}  SimViT-Large  &   62.9  &  12.2   & \textbf{83.4} \\ \hline
\end{tabular}}
\label{table:ImageClassificationCompareSOTA}
\vspace{-0.5cm}
\end{table}

\begin{table}[!h]
\centering
\caption{\small COCO object detection results of RetinaNet\cite{RetinaNet}, ATSS\cite{ATSS} and GFL\cite{GFL} based on our SimViT-Small model.}
\vspace{0.1cm}
\setlength{\tabcolsep}{0.3mm}{
\begin{tabular}{llcc|ccc}
\hline
Method                                             & \multicolumn{1}{c}{Backbone} & \#Param & FLOPs & 
\multicolumn{1}{l}{$AP^b$} & \multicolumn{1}{l}{$AP^b_{50}$} & \multicolumn{1}{l}{$AP^b_{75}$} \\ \hline

\multicolumn{1}{c}{\multirow{3}{*}{RetinaNet\cite{RetinaNet}}} & ResNet-50   & 37.7   & 239 & 39.0  & 58.4    & 41.8   \\
\multicolumn{1}{c}{}                               & Swin-T & 38.5   & 245  & 45.0 &  65.9  &  48.4 \\
\rowcolor{gray!20} \multicolumn{1}{c}{}                          & SimViT-Small & 39.1  & 281 & \textbf{46.3} & \textbf{67.5} & \textbf{49.6}  \\ \hline

\multicolumn{1}{c}{\multirow{3}{*}{ATSS\cite{ATSS}}}          & ResNet-50  & 32.1   & 205    & 43.5  & 61.9   & 47.0 \\
\multicolumn{1}{c}{}                               & Swin-T & 35.7   & 212   & 47.2  & 66.5   & 51.3  \\
\rowcolor{gray!20}  & SimViT-Small & 37.0  & 248  & \textbf{49.6}  & \textbf{68.9}   & \textbf{54.1} \\ \hline
\multicolumn{1}{c}{\multirow{3}{*}{GFL\cite{GFL}}}                       & ResNet-50 & 32.0   & 208   & 44.5  &  63.0 &  48.3 \\
                                                   & Swin-T  & 36.0   & 215   & 47.6 & 66.8 & 51.7 \\
         \rowcolor{gray!20}       & SimViT-Small & 37.1 & 251 & \textbf{49.9} & \textbf{69.3} & \textbf{54.1} \\ 
\hline
\end{tabular}}
\label{table:detection2}
\vspace{-0.2cm}
\end{table}

\subsection{Object Detection}
Our object detection tasks are based on COCO 2017 dataset\cite{coco}, which consists of 118K training and 5K validation images. The pretrained SimViT-Small backbone on ImageNet-1k will be further finetuned on various object detection models, e.g., RetinaNet\cite{RetinaNet}, ATSS\cite{ATSS} and GFL\cite{GFL}. The experimental results show our backbone model is compatible with different object detection methods. Following Swin\cite{Swin}, our models are trained for 36 epochs and be applied with a multi-scale training strategy. Results of all detectors are reported on the validation set and more details can be found in Table \ref{table:detection2}. Our SimViT-Small architecture brings consistent $+1.3\sim2.4$ box AP gains over Swin-T. 


\subsection{Semantic Segmentation}
We choose ADE20K\cite{ADE20K} as our benchmark dataset for semantic segmentation tasks. ADE20K contains more than 20K scene-centric images exhaustively annotated with pixel-level objects and object parts labels. We follow PVT\cite{PVT} to employ classical Semantic FPN\cite{SemanticFPN} as our segmentation framework. We use different-size pretrained SimViT models to finetune in the semantic segmentation task. We report our experimental results in Table \ref{table:SegmentationFPN}. Compared with ResNet\cite{ResNet} and PVT\cite{PVT}, our models bring $+5.7\% \sim 10.5\%$ mIOU gains.
\begin{table}[h]
\vspace{-0.5cm}
\caption{\small The semantic segmentation results of Semantic FPN on the ADE20K validation set. ``GFLOPS" is computed under the input size of 512x512.} 
\vspace{0.1cm}
\centering
\setlength{\tabcolsep}{1mm}{
\begin{tabular}{lccc}
\hline
\multirow{2}{*}{Backbone} & \multicolumn{3}{c}{Semantic FPN} \\ \cline{2-4} 
        &      Param(M) &   GFLOPs    &  mIOU(\%)     \\ \hline
        
        \rowcolor{gray!20}  SimViT-Micro  &    7.2   &   24.7   &   \textbf{37.5}    \\ \hline
      ResNet18\cite{ResNet}  &  15.5   & 32.2  & 32.9      \\ 
      PVT-Tiny\cite{PVT}  &   17.0    & 33.2  & 35.7  \\  
\rowcolor{gray!20}  SimViT-Tiny  &  16.8 &  36.1 &   \textbf{42.7}   \\ \hline

  ResNet50\cite{ResNet}    &    28.5   &    45.6   & 36.7   \\ 
      PVT-Small\cite{PVT}  &    28.2  & 44.5      & 39.8  \\
\rowcolor{gray!20}  SimViT-Small  &  33.2  &  54.0 & \textbf{47.2} \\ \hline

 ResNet101\cite{ResNet}    &    47.5   &    65.1   & 38.8  \\ 
      PVT-Large\cite{PVT}  & 65.1  &    79.6  & 42.1  \\  
      \rowcolor{gray!20}  SimViT-Medium  &   55.0  &  78.5    &   \textbf{47.8}   \\ \hline
\end{tabular}}
\label{table:SegmentationFPN}
\vspace{-0.5cm}
\end{table}

\subsection{Ablation Study}
\textbf{The effect of model depth.} Our similar-size models have more layers compared with PVT\cite{PVT} because of the high-efficient parameters of our models. In order to rule out the possibility of model depth causing our model to be superior to competitors, we conduct an ablation study in the situation of the same depth settings based on our SimViT-Tiny. Results in Table \ref{table:The effect of model depth} show that even fewer parameters(11.1M vs. 13.2M) than PVT\cite{PVT} in the same depth setting, the classification performance of our model is 2.7\% higher than PVT(77.8\% vs. 75.1\%). The performance of our model is further boosted along with deepening layers( 79.3\% vs. 77.8\%).\\

\begin{table}[htbp]
\vspace{-0.7cm}
\caption{\small Ablation study about the effect of model depth compared with PVT-Tiny\cite{PVT}. The experimental results are based on our SimViT-Tiny.} 
\vspace{0.2cm}
\centering
\setlength{\tabcolsep}{0.5mm}{
\begin{tabular}{ccccc}
\hline
   Method   & Depth   &  Param(M) &   GFLOPs    &  Top-1 Acc(\%)     \\ \hline
  PVT-Tiny \cite{PVT} & 2-2-2-2 & 13.2 & 1.9 & 75.1 \\  
SimViT(ours) &  2-2-2-2 &  11.1  & 2.0 &   77.8  \\ 
SimViT(ours) &  2-4-3-2 &  13.0  & 2.5 &   \textbf{79.3} \\ \hline
\end{tabular}}
\label{table:The effect of model depth}
\vspace{-0.1cm}
\end{table}

\begin{table}[htbp]
\caption{\small Ablation study on position embedding and the size of sliding windows for SimViT-Micro. `Pos. Embed' suggests that the model with absolute position embedding from PVT\cite{PVT}. `Window size' denotes that the size of the sliding windows in our MCSA of the first three stages.}
\vspace{0.2cm}
\label{tab:ablation2}
\centering
\setlength{\tabcolsep}{1.0mm}{
\begin{tabular}{c|c|c|c|c}
\hline
Pos. Embed                      & Window size & Param(M) & GFLOPs & Top-1(\%) \\ \hline
\checkmark & 3-3-3       & 3.4  & 0.65  & 70.97     \\ 
-                         & 5-5-5       & 3.3  & 0.65  & 69.89    \\ 
-                         & 3-3-3       & 3.3  & 0.65  & \textbf{71.08}      \\ \hline 
\end{tabular}}
\vspace{-0.4cm}
\end{table}

\textbf{The effect of sliding windows and position encoding.} We experimentally investigate the effect of window size and position encoding on the performance of our SimViT in Table \ref{tab:ablation2}. The experimental results show the larger window size does not result in improved results(69.89\% vs. 71.08\%). Position encoding is also not required in our SimViT(70.97\% vs. 71.08\%), this shows our simViT can naturally capture spatial structure without any explicit position encoding. The removal of position encoding also brings the translation invariance, which is important for the recognition ability of vision models.


\section{Conclusion}
In this paper, we propose a simple yet effective vision Transformer named SimViT, which integrates spatial structure and local information into vision Transformers using Multi-head Central Self-Attention(MCSA) along with simple sliding windows. Meanwhile, the multi-scale hierarchical features from our model can be applied to various dense prediction visual tasks. Extensive experiments  show the effectiveness of our SimViT as a general backbone model on image classification, object detection and semantic segmentation.  We expect our work can be applied to more vision tasks in the future.
\bibliographystyle{IEEEbib}
\bibliography{icme2022template}

\end{document}